\documentclass[11pt,a4paper]{article}
\usepackage[hyperref]{acl2021}
\usepackage{times}
\usepackage{latexsym}
\usepackage{hyperref}
\usepackage{multirow}
\usepackage{threeparttable}
\usepackage{comment}
\usepackage{graphicx}
\usepackage{makecell}
\usepackage{comment}

\usepackage{microtype}

\aclfinalcopy % Uncomment this line for the final submission
\title{UPB at SemEval-2021 Task 1: Combining Deep Learning and Hand-Crafted Features for Lexical Complexity Prediction}

\author{George-Eduard Zaharia, Dumitru-Clementin Cercel, Mihai Dascalu \\
  University Politehnica of Bucharest, Faculty of Automatic Control and Computers\\
  \tt  george.zaharia0806@stud.acs.upb.ro\\
  \tt \{dumitru.cercel, mihai.dascalu\}@upb.ro\\
 }
\date{}

\begin{document}
\maketitle
\begin{abstract}
Reading is a complex process which requires proper understanding of texts in order to create coherent mental representations. However, comprehension problems may arise due to hard-to-understand sections, which can prove troublesome for readers, while accounting for their specific language skills. As such, steps towards simplifying these sections can be performed, by accurately identifying and evaluating difficult structures. In this paper, we describe our approach for the SemEval-2021 Task 1: Lexical Complexity Prediction competition that consists of a mixture of advanced NLP techniques, namely Transformer-based language models, pre-trained word embeddings, Graph Convolutional Networks, Capsule Networks, as well as a series of hand-crafted textual complexity features. Our models are applicable on both subtasks and achieve good performance results, with a MAE below 0.07 and a Person correlation of .73 for single word identification, as well as a MAE below 0.08 and a Person correlation of .79 for multiple word targets. Our results are just 5.46\% and 6.5\% lower than the top scores obtained in the competition on the first and the second subtasks, respectively.
\end{abstract}

\section{Introduction}

Reading is a complex process due to the mental exercise readers are challenged to perform, since a coherent representation of the text needs to be projected into their mind in order to grasp the underlying content~\cite{articleBroek}. For non-native speakers, the lack of text understanding hinders knowledge assimilation, thus becoming the main obstacle that readers need to overcome. Complex words can impose serious difficulties, considering that their meaning is often dependant on their context and cannot be easily inferred. In order to facilitate text understanding or to perform text simplification, complex tokens first need to be detected. This can be performed by developing systems capable of identifying them by individual analysis, as well as contextual analysis. 

There are two main approaches regarding the complexity task.
Tokens can be binary classified as complex or non-complex, a procedure that helps users separate problematic words from the others. Words can be also labeled with a probabilistic complexity value, which in return can be used to simplify the text. Words with lower degrees of complexity can be easily explained, whereas more complex tokens can be replaced with simpler equivalent concepts.

The Lexical Complexity Prediction (LCP) shared task, organized as the SemEval-2021 Task 1 \cite{shardlow2021semeval}, challenged the research community to develop robust systems that identify the complexity of a token, given its context. Systems were required to be easily adaptable, considering that the dataset entries originated from multiple domains. At the same time, the target structure evaluated in terms of complexity could contain a single word or multiple words, depending on the subtask. 

The current work is structured as follows. The next section presents the state-of-the-art Natural Language Processing (NLP) approaches for LCP (probabilistic) and complex word identification (CWI). The third section outlines our approaches for this challenge, while the fourth section presents the results. Afterwards, the final section draws the conclusions and includes potential solutions that can be used to further improve performance.

\section{Related Work}
\textbf{Probabilistic CWI.} \newcite{kajiwarakomachi2018complex} adopted for the CWI task a system based on Random Forest regressors, alongside several features, such as the presence of the target word in certain corpora. Moreover, they conducted experiments to determine the best parameters for their regression algorithms. 

\newcite{dehertogtack2018deep} introduced a deep learning architecture for probabilistic CWI. Apart from the features extracted by the first layers of the network, the authors also included a series of hand-crafted features, such as psychological measures or frequency counts. Their architecture included different Long Short-Term Memory (LSTM) modules \cite{hochreiter1997long} for the input levels (i.e., word, sentence), as well as the previously mentioned psychological measures and corpus counts.

\textbf{Sequence labeling CWI.} \newcite{goodingkochmar2019complex} introduced a technique based on LSTMs for CWI, which obtained better results on their sequence labeling task than previous approaches based only on feature engineering. The contexts detected by the LSTM offered valuable information, useful for identifying complex tokens placed in sequences. 

Changing the focus towards text analysis, \newcite{finnimore2019strong} extracted a series of relevant features that supports the detection of complex words. While considering their feature analysis process, the greatest influence on the overall system performance was achieved by the number of syllables and the number of punctuation marks accompanying the targeted tokens.

A different approach regarding CWI was adopted by \newcite{zampierietal2017complex}, who employed the usage of an ensemble created on the top systems from the SemEval CWI 2016 competition \cite{paetzold2016semeval}. Other experiments performed by the authors also included plurality voting~\cite{polikar_robi}, or a technique named Oracle~\cite{kuncheva}, that forced label assignation only when at least one classifier detected the correct label.

\newcite{zaharia2020crosslingual} tackled CWI through a cross-lingual approach. Resource scarcity is simulated by training on a small number of examples from a language and testing on different languages, through zero-shot, one-shot, and few-shot scenarios. Transformer-based models \cite{vaswani2017attention} achieved good performance on the target languages, even though the number of training entries was extremely reduced.

\section{Method}

\subsection{Dataset}
CompLex~\cite{shardlow2020complex, shardlow2021predicting} is the dataset used for the LCP shared task that was initially annotated on a 5-point Likert scale. Moreover, the authors performed a mapping between the annotations and values between 0 and 1 in order to ensure normalization. The dataset has two categories, one developed for single word complexity score prediction, while the other is centered on groups of words; each category has entries for training, trial, and testing. The single word dataset contains 7,662 entries for training, 421 trial entries, and 917 test entries. The multi-word dataset contains a considerably smaller number of entries for each category, namely 1,517 for training, 99  trial entries, and 184 for testing.

All entries from the LCP shared task are part of one of three different English corpora (i.e., Bible - biblical entries, Biomed - biomedical entries, and Europarl - political entries), evenly distributed, each one representing approximately 33\% of its corresponding set. As such, the task is even more challenging when considering the vastly different domains of these entries.

\begin{comment}
\begin{table*}[!htb]
\small
\centering
\caption{Entry distribution in the LCP 2021 dataset.}
\begin{threeparttable}
\begin{tabular}{|c|c|c|c|}\hline 
\multicolumn{2}{|c|}{\bf{Dataset}} & \bf Single  & \bf Multi \\
\hline
\multirow{4}{*}{Training} & Bible & 2574 & 505 \\
 & Biomed & 2576 & 514 \\
  & Europarl & 2512 & 498 \\
  \cline{2-4}
  & Total & 7662 & 1517 \\
   \hline 
\multirow{4}{*}{Trial} & Bible & 143 & 29 \\
 & Biomed & 135 & 33 \\
  & Europarl & 143 & 37 \\
  \cline{2-4}
  & Total & 421 & 99 \\
  \hline
  \multirow{4}{*}{Testing} & Bible & 283 & 66 \\
 & Biomed & 289 & 53 \\
  & Europarl & 345 & 65 \\
  \cline{2-4}
  & Total & 917 & 184\\
 \hline
\end{tabular}
\end{threeparttable}
\label{tab:tablelcp} 
\end{table*}
\end{comment}

\subsection{Architecture}
\begin{figure*}[h!]
\centering
\includegraphics[width=0.7\linewidth]{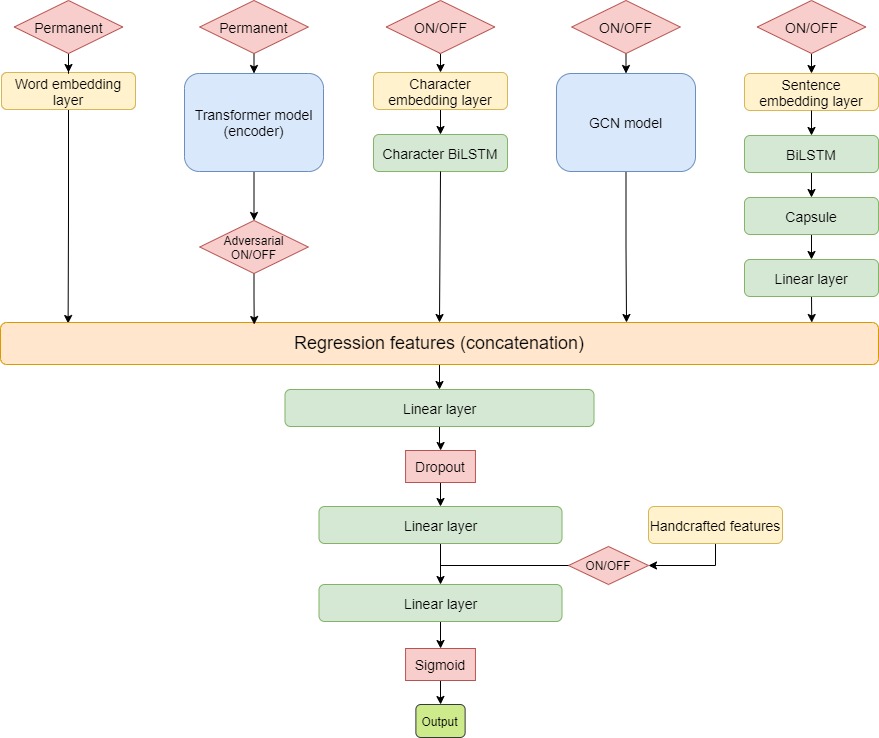}
  \caption{The overall model architecture used in our experiments.}
  \label{fig:finalmodel}
\end{figure*}

During our experiments, we combined features obtained from multiple modules described later on in detail, and then applied three regression layers, alongside a \textit{Dropout} layer, to obtain the complexity score of the input (see Figure \ref{fig:finalmodel} for our modular architecture). The permanent components are represented by the target word embeddings and the Transformer features, which are concatenated and then fed into the final linear layers, designated for regression. The other components (i.e., character-level embeddings, GCN, and Capsule) are enabled in particular setups; similarly, the adversarial training component can also be disabled. At the same time, a series of hand-crafted features can be concatenated before the last layer with the aim to further improve the overall performance.

\subsection{Pre-trained Word Embeddings}
Pre-trained word embeddings were used as features for the final regression as an initial representation of the input. Throughout our experiments, three types of pre-trained word embeddings were considered, namely: GloVe\footnote{\url{https://nlp.stanford.edu/projects/glove/}}, FastText\footnote{\url{https://fasttext.cc/}}, and skip-gram\footnote{\url{http://vectors.nlpl.eu/repository/}}. Out of the three previous options, GloVe performed best in our experiments. As such, the results section exclusively reports the performance obtained by our configurations alongside GloVe embeddings for the target word.

\subsection{Transformer-based Language Models}
Considering that Transformers achieve state-of-the-art performance for most NLP tasks~\cite{wolfetal2020transformers}, all our setups include a Transformer-based component. However, they are pre-trained in different manners; thus, we experimented with several variants, as follows:

\begin{itemize}
\item \textbf{BERT}~\cite{devlin2019bert} - Extensively pre-trained on  English, BERT-base represents the baseline of Transformer-based models;
\item \textbf{BioBERT}~\cite{10.1093/bioinformatics/btz682} - Considering that some of the most difficult to understand entries are part of the Biomed corpus, we also experimented with a model pre-trained on biomedical data;
\item \textbf{SciBERT}~\cite{Beltagy2019SciBERT} - Similarly to BioBERT, SciBERT is pre-trained on scientific data and becomes a good candidate for fine-tuning on the scientific entries from the dataset;
\item \textbf{RoBERTa}~\cite{liu2019roberta} - RoBERTa improves upon BERT by modifying key hyperparameters, and by being trained with larger mini-batches and learning rates; RoBERTa usually has better performance on downstream tasks.
\end{itemize}

\subsection{Adversarial Training}
We also aimed to improve the robustness of the main element of our architecture, the Transformer-based component. Therefore, we adopted an adversarial training technique, similar to \newcite{karimi2020adversarial}. The adversarial examples generated during training work on the embeddings level, and are based on a technique that uses the gradient of the loss function.

\subsection{Character-level Embeddings}
Alongside the previously mentioned word embeddings for the target word, we also employ character-level embeddings for the same word, such that its internal structure, as well as its universal context, can be properly captured as features in our architecture.

\subsection{Graph Convolutional Networks}
Besides the previous Transformer-based models, we also explored the relations between the dataset entries, as well as the vocabulary words. Therefore, Graph Convolutional Networks (GCN) \cite{kipf2017semisupervised} were also considered for determining node embedding vectors, by taking into account the properties of the neighboring nodes. By stacking multiple GCN layers, the information embedded into a node can become broader, inasmuch as it incorporates considerably larger neighborhoods. Similar to \newcite{yao2018graph}, we consider the graph to have several nodes equal to the number of entries (documents) in the corpus plus the vocabulary size of the corpus.

\subsection{Capsule Networks}
Alongside the relational approach derived from GCN and Transformer embeddings, we intended to further analyze our inputs by passing them through a Capsule Network~\cite{sabour2017dynamic}. This approach enables us to obtain features that reflect aspects specific to different levels of the inputs, as Capsule Networks increase the specificity of features, while the capsule layers go deeper.

\subsection{Hand-crafted Features}
Similar to \newcite{goodingkochmar2018camb}, we integrated a series of hand-crafted features for the target word: \textbf{Syllables}, \textbf{Synset length}, \textbf{Hypernyms length}, \textbf{Hyponyms length}, \textbf{Number of dependencies} obtained using NLTK\footnote{\url{https://www.nltk.org/}} alongside CoreNLP\footnote{\url{https://stanfordnlp.github.io/CoreNLP/}}, \textbf{SubIMDB presence}\footnote{\url{http://ghpaetzold.github.io/subimdb/}}, \textbf{SimpWiki presence}~\cite{costersimpwiki}, \textbf{CEFR level}  obtained from the Cambridge English dictionary\footnote{\url{https://dictionary.cambridge.org/dictionary/learner-english/}}, \textbf{MRC features} \cite{mrcfeatures} (\textit{Age of acquisition}, \textit{Concreteness rating}, \textit{Imageability rating}, \textit{Word familiarity rating }, \textit{Number of phonemes}), \textbf{Semantic Diversity}~\cite{semanticdiversity}, \textbf{Sensorimotor Norms}~\cite{lancastersensorimotor}.

\textbf{Character n-grams} - The character n-gram approach consists of two steps: first, a vectorizer is applied on the inputs to select a maximum number of 5,000 most frequent n-grams; second, Tf-Idf scores for these elements are computed. The obtained values are then normalized in the $ [0,1] $ range and used as features.

\textbf{ReaderBench indices} - The ReaderBench framework~\cite{dascalu2017readerbench} was used to extract additional textual complexity features reflective of writing style. Out of the 1311 features obtained by running ReaderBench on our inputs, we selected 278. The choice was made by considering only the features with a high linguistic coverage (i.e, were non-zero for at least 50\% of the entries).

\subsection{Traditional Machine Learning Baseline}
Several machine learning algorithms, such as \textit{Logistic regression}, \textit{Random Forest Regressors}, \textit{XGBoost regression}, or \textit{Ridge regression} were experimented using the aforementioned handcrafted features.

We then switched to a ridge regression approach and trained it with a multitude of features, including Transformer-based embeddings (BERT, BioBERT, SciBERT, RoBERTa), pre-trained word embeddings (GloVe, fastText, Skip-gram), and handcrafted features. 

\subsection{Preprocessing and Experimental Setup}
Text preprocessing is minimal and consists of removing unnecessary punctuation, such as quotes. The experimental hyperparameters for all modules are presented in Table \ref{tab:lcp_hyper}.

\begin{table*}[ht]
\small
\begin{center}
\caption{Experimental hyperparameters for the probabilistic CWI.}
\begin{tabular}{|l|l|l|l|l|l|}
\hline \bf Transformers & \bf GCN & \bf Capsule & \bf BiLSTM & \bf Embedding & \bf Full Model \\ \hline
Size: 768 & \makecell[l]{GCN Size 1: 512 \\ GCN Size 2: 256} & \makecell[l]{Routings: 5 \\ Number of \\ capsules: 10 \\ Capsule \\ dimension: 16} & Dimension: 128 & Dimension: 300 & \makecell[l]{Optimizer: AdamW \\ Loss Function: MSELoss \\ Learning Rate: \textit{2e-5}} \\ \hline
\end{tabular}
\label{tab:lcp_hyper}
\end{center}
\end{table*}

\section{Results}
Table \ref{tab:lcpdeep} introduces the results obtained using our deep learning architecture, while Table \ref{tab:lcpml} focuses on the traditional machine learning baseline. The best results for the deep learning approaches applied on the single target word dataset are obtained using RoBERTa as Transformer model. The setup which maximizes performance considers RoBERTa, GCN, and Capsule features, obtaining a Pearson score of 0.7702 and a mean absolute error (MAE) of 0.0671 on the trial dataset. Moreover, the high performance is maintained on the test dataset, with a Pearson correlation coefficient of 0.7237 and a MAE of 0.0677. BERT, SciBERT, and BioBERT have similar results with marginal differences; GCN, Capsule, and adversarial training improve performance for all models, while character-level embeddings do not provide a boost in performance.

\begin{table*}[!htb]
\small
\centering
\caption{Results for the Deep Learning approaches.}
\begin{threeparttable}
\begin{tabular}{|l|c|c|c|c|c|c|c|c|}
\hline 
\multirow{3}{*}{\textbf{Configuration}} & \multicolumn{4}{c|}{\bf{Single-Word Target}} & \multicolumn{4}{c|}{\bf{Multi-Word Target}} \\ \cline{2-9}
& \multicolumn{2}{c|}{\bf{Trial}} & \multicolumn{2}{c|}{\bf{Test}} & \multicolumn{2}{c|}{\bf{Trial}} & \multicolumn{2}{c|}{\bf{Test}} \\ \cline{2-9}
& \bf Pearson & \bf MAE & \bf Pearson & \bf MAE & \bf Pearson & \bf MAE & \bf Pearson & \bf MAE  \\ \hline
BERT & 0.7575 & 0.0689  & 0.7170 & 0.0682 & 0.7019  & 0.0969 & 0.7853 & 0.0781  \\ \hline
BERT + Capsule & 0.7641 & 0.0685 & 0.7113 & 0.0689 & 0.7170 & 0.0958 & 0.7774 & 0.0791 \\ \hline
BERT + GCN + Capsule & 0.7548 & 0.0693 & 0.7178 & 0.0682 & 0.6978 & 0.0905 & 0.7773 & 0.0812 \\ \hline
\makecell[l]{BERT + GCN + Capsule \\ + Adversarial Training} & 0.7608 & 0.0695 & 0.7171 & 0.0684 & 0.7077 & 0.0933 & 0.8008 & 0.0779\\  \hline
BERT + Char Embeddings & 0.7505 & 0.0701 & 0.6925 & 0.0717 & 0.7091 & 0.0904 & 0.7800 & 0.0821  \\ 
\hline
\hline
RoBERTa & 0.7676 & 0.0685 & 0.7222 & 0.0681 & 0.7177 & 0.0925 & 0.7921 & 0.0764 \\ \hline
RoBERTa + GCN + Capsule\textit{*} & \textbf{0.7702} & \textbf{0.0671} & 0.7237 & \textbf{0.0677} & 0.7160 & 0.0910 & \textbf{0.7962} & 0.0788 \\ \hline
\makecell[l]{RoBERTa + GCN + Capsule \\ + Adversarial Training\textit{*}} & 0.7699 & 0.0682 & \textbf{0.7324} & 0.0703 & \textbf{0.7227} & 0.0893 & 0.7851 & 0.0808  \\ \hline
\makecell[l]{RoBERTa + \\ Hand-crafted Features} & 0.7476 & 0.0704 & 0.7028 & 0.0735 & 0.7165 & 0.0974 & 0.7932 & \textbf{0.0754} \\ \hline
\makecell[l]{RoBERTa + Char Embeddings \\ + Capsule + GCN + Adversarial} & 0.7663 & 0.0696 & 0.7264 & 0.0692 & 0.7221 & 0.0954 & 0.7958 & 0.0791 \\ \hline
RoBERTa + Char Embeddings & 0.7658 & 0.0695 & 0.7259 & 0.0682 & 0.7167 & 0.0958 & 0.7916 & 0.0772 \\ 
\hline
\hline
SciBERT + GCN & 0.7626 & 0.0714 & 0.7145 & 0.0715 & 0.6829 & 0.0876 & 0.7888 & 0.0762 \\ \hline
\makecell[l]{SciBERT + GCN + Capsule \\ + Adversarial Training} & 0.7617 & 0.0721 & 0.7086 & 0.0724 & 0.7164 & \textbf{0.0863} & 0.7882 & 0.0785 \\ \hline
SciBERT + Char Embeddings & 0.7512 & 0.0710 & 0.7079 & 0.0691 & 0.6855 & 0.1046 & 0.7729 & 0.0809 \\ 
\hline
\hline
BioBERT & 0.7658 & 0.0694 & 0.7151 & 0.0698 & 0.7014 & 0.0906 & 0.7814 & 0.0827 \\ \hline
\makecell[l]{BioBERT + GCN + Capsule \\ + Adversarial Training} & 0.7683 & 0.0677 & 0.7144 & 0.0690 & 0.7098 & 0.0968 & 0.7919 & 0.0795 \\ \hline
BioBERT + Char Embeddings & 0.7619 & 0.0689 & 0.7073 & 0.0697 & 0.7069 & 0.0995 & 0.7849 & 0.0810 \\
\hline
\end{tabular}
\begin{tablenotes}\footnotesize
\item[*] The models marked with \textit{*} are the ones used in our submissions.\\
\end{tablenotes}
\end{threeparttable}
\label{tab:lcpdeep}
\end{table*}

Table \ref{tab:lcpml} presents the results obtained using the features described in Section 3, namely Transformer-based contextualized embeddings (BERT, RoBERTa, BioBERT, SciBERT), pre-trained word embeddings (GloVe, fastText, skip-gram), and hand-crafted features, all combined using various regression algorithms. Logistic regression, Random Forrest and XGBosst yield lower performance when compared to the previous deep learning approaches. However, we managed to increase the scores on the single target word dataset, with Pearson coefficients of 0.7738 and 0.7340 on the trial and test datasets, by combining the results obtained from training several instances of ridge regression. Nevertheless, the best results for the multiple target word task are still obtained by the deep learning approaches (RoBERTa, GCN, Capsule, adversarial training), which surpass the Ridge Regression + pre-trained word embeddings + Transformer embeddings + handcrafted features approach by a low margin of 0.0074 Pearson on the trial dataset and 0.0033 on the test dataset.

\begin{table*}[!htb]
\small
\centering
\caption{Results for the Traditional Machine Learning solutions.}
\begin{threeparttable}
\begin{tabular}{|l|c|c|c|c|c|c|c|c|}
\hline 
\multirow{3}{*}{\textbf{Method}} & \multicolumn{4}{c|}{\bf{Single-Word Target}} & \multicolumn{4}{c|}{\bf{Multi-Word Target}} \\ \cline{2-9}
& \multicolumn{2}{c|}{\bf{Trial}} & \multicolumn{2}{c|}{\bf{Test}} & \multicolumn{2}{c|}{\bf{Trial}} & \multicolumn{2}{c|}{\bf{Test}} \\ \cline{2-9}
& \bf Pearson & \bf MAE & \bf Pearson & \bf MAE & \bf Pearson & \bf MAE & \bf Pearson & \bf MAE  \\ \hline

Logistic Regression & 0.7158 & 0.0748 & 0.6868 & 0.0718 & 0.6533 & 0.0954 & 0.7558 & 0.0791 \\ \hline
\makecell[l]{Random Forest Regressor} & 0.7390 & 0.0708 & 0.7011 & \textbf{0.0691} & 0.6714 & 0.0929 & 0.7651 & \textbf{0.0785} \\ \hline
XGBoost Regressor & 0.7488 & 0.0700 & 0.7033 & 0.0695 & 0.6503 & 0.0975 & 0.7544 & 0.0804 \\ \hline
\makecell[l]{Ridge Regression\textit{*}} & \textbf{0.7738} & \textbf{0.0686} & \textbf{0.7340} & 0.0699 & \textbf{0.7153} & \textbf{0.0873} & \textbf{0.7929} & 0.0787 \\ \hline
\end{tabular}
\begin{tablenotes}\footnotesize
\item[*] The solution marked with \textit{*} is the one used in our submissions.\\
\end{tablenotes}
\end{threeparttable}
\label{tab:lcpml}
\end{table*}

\section{Discussion}

\begin{table*}[!ht]
\small
\begin{center}
\caption{Difficult Biomed entries.}
\begin{tabular}{|l|c|c|c|}
\hline \bf Entry & \bf Target & \bf Predicted complexity & \bf True complexity \\ \hline
\makecell[l]{Genetic analyses of sitosterolemia pedigrees allowed the \\ mapping  of the STSL locus to human chromosome 2p21, \\ between D2S2294 and D2S2298 [12,13].} & pedigrees & 0.4516 & 0.3125 \\ \hline
\makecell[l]{Normally cells accumulate H3-2meK9 and H3-3meK9 \\ marks and HP1B protein on the sex chromatin as they \\ progress to diplonema, but we observed mutant diplotene \\ cells lacking these features.} & marks  & 0.2686  & 0.3409 \\ \hline
\makecell[l]{p150CAF-1 knockdown in ES cells was quantified by \\ Western blot analysis and IF.} & ES  & 0.5587  & 0.6944 \\ \hline
\end{tabular}
\label{tab:tabbiomed}
\end{center}
\end{table*}

The entries with the largest difference when compared to the gold standard are represented by the ones that are part of the Biomed category. This discrepancy is valid for both subtasks (i.e, single target word and multiple target words). The Biomed entries employ the usage of more complex terminology, quantities, or specific scientific names. Therefore, it becomes more difficult for standard pre-trained Transformer systems, such as BERT or RoBERTa, to adapt to the Biomed entries. In contrast, corpora with easier to understand language (i.e., Bible and Europarl) are not properly represented when using BioBERT or SciBERT, considering that the Transformers are mainly pre-trained for scientific or biomedical texts. 

Moreover, a considerable part of the Biomed entries contains large amounts of abbreviations, while other entries from the same domain have specific terms or links, as seen in Table \ref{tab:tabbiomed}. The difference between our predictions and the correct labels are up to 0.14 for the complexity probability.

\section{Conclusions and Future Work}
This work proposes a modular architecture, as well as different training techniques for the Lexical Complexity Prediction shared task. We experimented with different variations of the previously mentioned architecture, as well as textual features alongside machine learning algorithms. First, we used different word embeddings and Transformer-based models as the main feature extractors and, at the same time, we examined a different training technique based on adversarial examples. Second, other different models were added, such as character-level embeddings, Graph Convolutional Networks, and Capsule Networks. Third, several hand-crafted features were also considered to create a solid baseline covering both deep learning and traditional machine learning regressors.

For future work, we intend to experiment with altering the modular architecture such that the models are trained similar to a Generative Adversarial Network~\cite{croceetal2020gan}, thus further improving robustness and achieving higher scores in terms of both Pearson correlation coefficients and MAE.

\bibliographystyle{acl_natbib}
\bibliography{anthology,acl2021}

%\appendix

\end{document}